\documentclass[11pt,a4paper]{article}
\usepackage[hyperref]{acl2017}
\usepackage{times}  
\usepackage{latexsym}
\usepackage{graphicx}
\usepackage{url}
\usepackage{multirow}
\usepackage{amsmath}
\usepackage{tikz}
\usepackage[T1]{fontenc}
\usepackage{dblfloatfix}

\aclfinalcopy 
\def\aclpaperid{165} 

\title{Domain Control for Neural Machine Translation}

\author{
 Catherine Kobus \ \  {\normalfont and} \ \  Josep Crego  \ \  {\normalfont and} \ \  Jean Senellart \\
  {\tt firstname.lastname@systrangroup.com} \\
  SYSTRAN / 5 rue Feydeau, 75002 Paris, France \\
\\}

\date{}

\begin{document}
\maketitle
\begin{abstract}
  Machine translation systems are very sensitive to the domains they were trained on. Several domain adaptation techniques have been deeply studied.
We propose a new technique for neural machine translation (NMT) that we call domain control which is performed at runtime using a unique neural network covering multiple domains. The presented approach shows quality improvements when compared to dedicated domains translating on any of the covered domains and even on out-of-domain data. In addition, model parameters do not need to be re-estimated for each domain, making this effective to real use cases.
Evaluation is carried out on English-to-French translation for two different testing scenarios. We first consider the case where an end-user performs translations on a known domain. Secondly, we consider the scenario where the domain is not known and predicted at the sentence level before translating. Results show consistent accuracy improvements for both conditions.
\end{abstract}

\begin{figure*}[h!]
\includegraphics[width=0.99\textwidth]{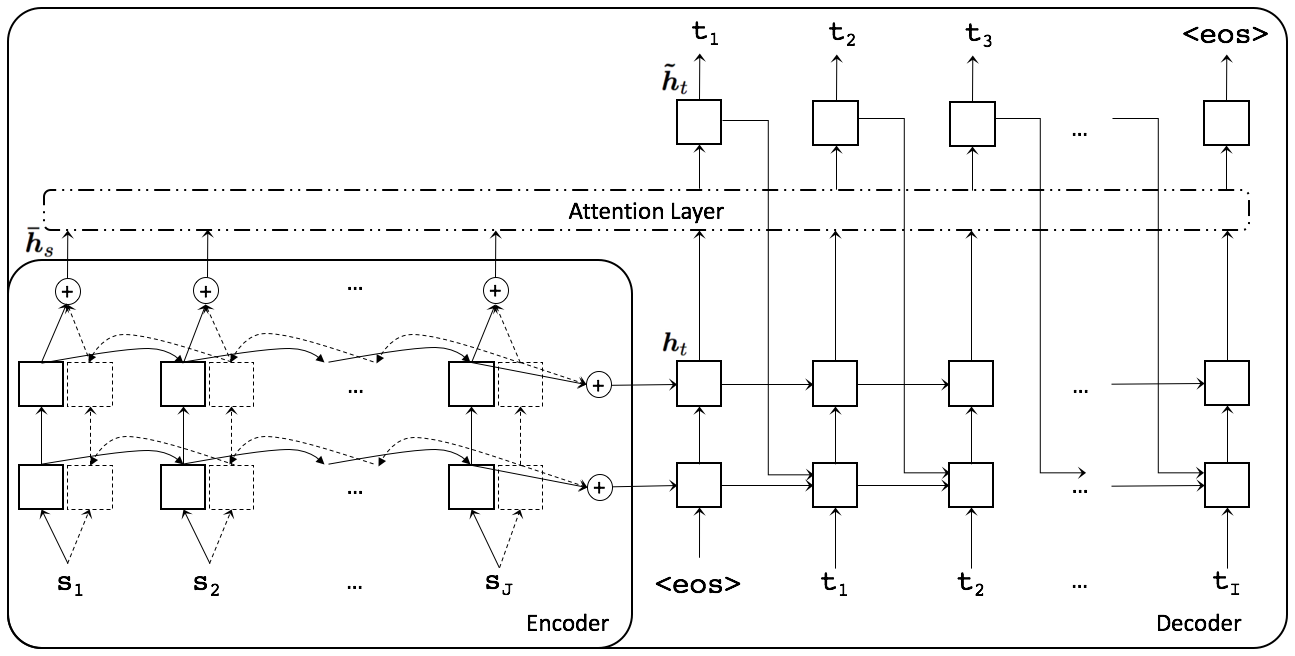}
\caption{Schematic view of our MT network.}
\label{opennmt}
\end{figure*}

\section{Introduction}
\label{sec:intro}

Machine translation systems are very sensitive to the domain(s) they were trained on because each domain has its own style, sentence structure and terminology. There is often a mismatch between the domain for which training data are available and the target domain of a machine translation system. If there is a strong deviation between training and testing data, translation quality will be dramatically deteriorated.
Word ambiguities are often an issue for machine translation systems. For instance, the English word \textit{"administer"} has to be translated differently if it appears in medical or political contexts.
Our work is motivated by the idea that neural models could benefit from having domain information to choose the most appropriate terminology and sentence structure while using the information from \textit{all} the domains to improve the base translation quality.
Recently, \cite{sennrich-haddow-birch_2016_NAACLHLT} report on the neural network ability to control politeness through side constraints. We extend this idea to domain control.
Our goal is to allow a model built from a diverse set of training data to produce in-domain translations. This is, to extend the coverage of generic NMT models to specific domains, with their specialized terminology and style, without lowering translation quality on more generic data. We present two frameworks to feed domain meta-information on the NMT encoder side.

The paper is structured as follows: Section \ref{sec:related} overviews related work. Details of our neural MT engine are given in Section \ref{sec:neural}. Section \ref{sec:domad} describes the proposed approach. Experiments and results are detailed in Section \ref{sec:exp}. Finally, conclusions and further work are drawn in Section \ref{sec:conclusions}.

\section{Related Work}
\label{sec:related}

A lot a work has already been done for domain adaptation in Statistical Machine Translation. The approaches vary from in-domain data selection based methods \cite{Hildebrand:2005:EAMT} \cite{Moore2010} \cite{sethy-georgiou-narayanan:2006:HLT-NAACL06-Short} to in-domain models mixture-based methods \cite{foster-kuhn:2007:WMT} \cite{koehn-schroeder:2007:WMT} \cite{Schwenk:2008:IJCNLP}.

Regarding neural MT \cite{Luong2015} adapt a generic NMT network (trained on out-of-domain data) by running additional training iterations over an in-domain data set. The authors claim to obtain a domain adapted model in a very limited training time. However, it differs from our work since we aim at performing domain-adapted translations using a unique network that covers multiple domains.


Recent works have especially dealt with domain adaptation for NMT by providing meta-information to the Neural Network. Our work is in line with this kind of approach. \cite{DBLP:journals/corr/ChenMKP16} feeds Neural Network with topic information on the decoder side; topics are numerous and consist in human-labeled product categories. \cite{topic_informed} includes topic modelling on both encoder and decoder sides. A given number of topics are automatically inferred from the training data using LDA; each word in a sentence is assigned its own vector of topics. In our work, we also provide meta-information about domain to the network. However, we introduce domain information at the sentence level.

\section{Neural MT}
\label{sec:neural}
Our NMT system follows the architecture presented in ~\cite{DBLP:journals/corr/BahdanauCB14}. It is implemented as an encoder-decoder network with multiple layers of a RNN with Long Short-Term Memory hidden units ~\cite{DBLP:journals/corr/ZarembaSV14}. Figure \ref{opennmt} illustrates an schematic view of the MT network.

Source words are first mapped to word vectors and then fed into a bidirectional recurrent neural network (RNN) that reads an input sequence $s = (s_1,...,s_J)$. Upon seeing the \texttt{<eos>} symbol, the final time step initialises a target RNN. The decoder is a RNN that predicts a target sequence $t = (t_1, ..., t_I)$, being $J$ and $I$ respectively the source and target sentence lengths. Translation is finished when the decoder predicts the \texttt{<eos>} symbol.

The left-hand side of the figure illustrates the bidirectional encoder, which actually consists of two independent LSTM encoders: one encoding the normal sequence (solid lines) that calculates a forward sequence of hidden states 
$(\overrightarrow{h_1}, ..., \overrightarrow{h_J})$, the second encoder reads the input sequence in reversed order (dotted lines) and calculates the backward sequence $(\overleftarrow{h_1},..., \overleftarrow{h_J})$. 
The final encoder outputs $(\overline{h}_1, ..., \overline{h}_J)$ consist of the sum of both encoders final outputs. 
The right-hand side of the figure illustrates the RNN decoder. Each word $t_i$ is predicted based on a recurrent hidden state $h_i$ and a context vector $c_i$ that aims at capturing relevant source-side information.

\begin{figure}[h!]
   \includegraphics[width=0.48\textwidth]{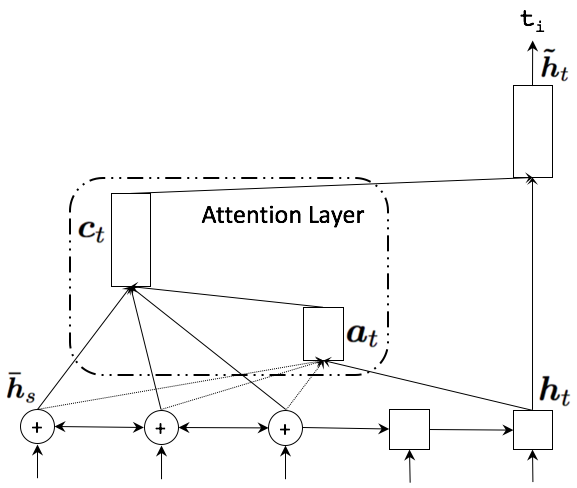}
\caption{Attention layer of the MT network.}
\label{attention}
\end{figure}

Figure \ref{attention} illustrates the attention layer. It implements the "general" attentional architecture from ~\cite{luong-pham-manning:2015:EMNLP}. 
The idea of a global attentional model is to consider all the hidden states of the encoder when deriving the context vector $c_t$.
Hence, global alignment weights $a_{t}$ are derived by comparing the current target hidden state $h_t$ with each source hidden state $\overline{h}_s$:

\begin{equation*}
a_{t}(s) = \frac{exp(score(h_t,\overline{h}_s))}{\sum_{s'} exp(score(h_t,\overline{h}_{s'}))}
\end{equation*}

with the content-based score function:
\begin{equation*}
score(h_t,\overline{h}_s) = h_t^T W_a \overline{h}_s
\end{equation*}

Given the alignment vector as weights, the context vector $c_t$ is computed as the weighted average over all the source hidden states.


The framework is available on the open-source project \texttt{seq2seq-attn}\footnote{\url{http://nlp.seas.harvard.edu}}.
More details about our system can be found in {\it Anonymised}. 

\section{Domain control}
\label{sec:domad}

Two different techniques are implemented to integrate domain control: additional token and domain feature.

\subsection{Additional Token}
\label{ssec:token}

The additional token method, inspired by the politeness control technique detailed in \cite{sennrich-haddow-birch_2016_NAACLHLT} consists in adding an artificial token at the end of each source sentence in order to let the network pay attention to the domain of each sentence pair. For instance, consider the next English-French translation: 

\begin{table}[h]
\begin{tabular}{ll}
Src: & { \small \texttt{Headache may be experienced}} \\
Tgt: & { \small \texttt{Des céphalées peuvent survenir}} \\
\end{tabular}
\end{table}

The network reads off the sentence pair with the appropriate \textit{Medical} domain tag \textbf{@MED@}:


\begin{table}[h]
\begin{tabular}{ll}
Src: & { \small \texttt{Headache may be experienced \textbf{@MED@}}} \\
Tgt: & { \small \texttt{Des céphalées peuvent survenir}} \\
\end{tabular}
\end{table}

Domain tags are appropriately selected in order to avoid overlaps with words present in the source language vocabulary. This method, though simple, has already proven to be effective to control the politeness level of a translation \cite{sennrich-haddow-birch_2016_NAACLHLT}, or to support multi-lingual NMT models \cite{zero_shot}.

\subsection{Word Feature}
\label{ssec:feature}

We present a second technique to introduce domain control in our neural translation model. We use word-level features as described in \cite{DBLP:journals/corr/CregoKKRYSABCDE16}. The first layer of the network is the word embedding layer. We adapt this layer to extend each word embedding with an arbitrary number of cells, designed to encode domain information. Notice that using additional features does not increase the vocabulary of source words; there are separate vocabularies for words and domain tags. Figure \ref{fig:feature} illustrates a word embedding layer extended with domain information. 

\begin{figure}[h]
   \includegraphics[width=0.48\textwidth]{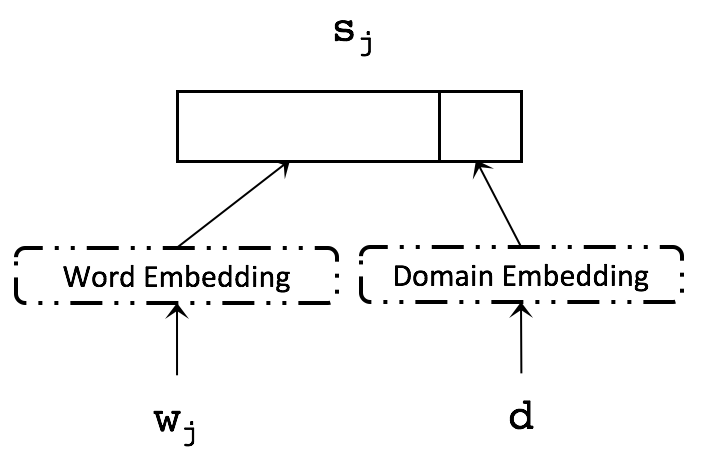}
\end{figure}
\label{fig:feature}

Following with the example of Section \ref{ssec:token}, the sentence pair is given to the network with the appropriate \textit{Medical} domain tag on each source word as follows:

\begin{flushleft}
\begin{table}[h!]
\begin{tabular}{ll}
Src: & {\small \texttt{Headache \ may \ be \ experienced}} \\
       & {\small \texttt{MED \ \ \ \ \ \ MED \ MED MED}} \\
Tgt: & {\small \texttt{Des céphalées peuvent survenir}} \\
\end{tabular}
\end{table}
\end{flushleft}

\begin{table*}[!h]
\begin{center}
\begin{tabular}{|lccc|ccc|}
\hline
Domain & Lines & Src words & Tgt words  & Lines & Src words & Tgt words\\
\hline
 & \multicolumn{3}{l}{Train} & \multicolumn{3}{l}{Test} \\
\hline
IT 	          & 399k & 6.0M  & 7.3M &   2k & 36,8k & 45,1k \\
Literature    & 35k  & 881k  & 943k & 2k & 50.1k & 54.0k \\
Medical       & 923k & 10.5M & 12.3M & 2k & 35.6k & 43.0k \\		  
News          & 194k & 5.4M  & 6.7M & 2k & 53.5k & 66.4k \\
Parliamentary & 1.6M & 37.6M & 43.8M & 2k & 40.7k & 49.4k \\
Tourism       & 1.1M & 23.3M & 27.5M & 2k & 39.1k & 45.5k \\
\hline
\bf Total     &\bf 4,3M & \bf 83,7M & \bf 98.5M & & &\\
\hline   
\end{tabular}
\end{center}
\caption{\label{tab:data} Statistics for training and test sets of each domain corpus. Note that k stand for thousands and M for millions.}
\end{table*}

Note that under this feature framework, the sentence-level domain information is added on a word-by-word basis to all the words in a sentence. We reuse an existing framework that was originally implemented to include linguistic features at the word level (Anonymised). 

\section{Experiments}
\label{sec:exp}

We evaluate the presented approach on English-to-French translation. Section \ref{ssec:training} describes the data used for the experiments and details training configurations. Finally, Section \ref{ssec:results} reports on translation accuracy results.

\subsection{Training Details}
\label{ssec:training}

We used training corpora covering six different domains : \textit{IT}, \textit{Literature}, \textit{Medical}, \textit{News}, \textit{Parliamentary} and \textit{Tourism}.  \textit{Medical}, \textit{News}, \textit{Parliamentary} data come from public corpora (respectively EMEA, News Commentary and Europarl), available from the OPUS repository \cite{Tiedemann2012parallel}. \textit{IT}, \textit{Literature} and \textit{Tourism} are proprietary data. Statistics of the corpora used are given in Table \ref{tab:data}.


\begin{table*}[h!]
\begin{center}
\begin{tabular}{|c|cc|cc|cc|}
\hline 
 \bf Domain & \bf Single & \bf Join & \bf Token & \bf Feature & \multicolumn{2}{c|}{\bf Feature} \\ 
 Constraint & \multicolumn{2}{c|}{None} & \multicolumn{2}{c|}{Oracle} & RNN & Acc (\%) \\ 
\hline
IT                    & 52.73 & 53.81 & 53.76 & {\bf ~54.56} (+0.75)  & 54.42 & 97.8 \\
Literature        & 20.25 & 29.81 & 29.96 & {\bf ~30.73} (+0.92) & 30.71 & 93.1 \\
Medical           & 33.97 & 41.83 & 42.02 & {\bf ~42.51} (+0.68) & 42.34 & 89.4 \\
News              & 29.70 & 33.83 & 34.47 & {\bf ~34.61} (+0.78)  & 34.49 & 88.3 \\
Parliamentary & 37.34 & 37.53 & 37.13 & {\bf ~37.79} (+0.26) & 37.77 & 82.7 \\
Tourism           & 37.05 & 37.46 & 37.72 & {\bf ~38.30} (+0.84) & 38.01 & 90.6 \\
\hline
Dialogs           & & 19.25 & &               & 19.55 & \\
\hline
\end{tabular}	
\end{center}
\caption{\label{tab:results} BLEU scores for the different systems and RNN-based domain classifier accuracy.}
\end{table*}

All experiments employ the NMT system detailed in Section \ref{sec:neural} and are performed on NVidia GeForce GTX 1080. We use BPE\footnote{\url{https://github.com/rsennrich/subword-nmt}} with a total of $32,000$ source and target tokens as vocabulary, computed over the entire training corpora. Word embedding size is $500$ cells. During training, we use stochastic gradient descent, a minibatch size of $64$ with dropout probability set to $0.3$ and bidirectional RNN. We train our models for $18$ epochs. Learning rate is set to $1$ and starts decaying after epoch $10$ by $0.5$. It takes about $10$ days to train models on the complete training data set ($4,3$M sentence pairs).

Four different training configurations are considered. The first includes six in-domain NMT models. Each model is trained using its corresponding domain data set (henceforth \textit{Single} models). The \textit{Join} network configuration is built using all the training data after concatenation. Note that this model does not include any information about domain. A \textit{Token} network is also trained using all the available training data. It includes domain information through the additional token approach detailed in Section \ref{ssec:token}. Finally, \textit{Feature} network is also trained on all available training data, it introduces domain information in the model by means of the feature framework detailed in Section \ref{ssec:feature}.

\subsection{Results}
\label{ssec:results}

\begin{table*}[h!]
\begin{center}
\begin{tabular}{|ll|}
\hline
Src: & Your doctor's instructions should be {\bf carefully observed} . \\
Ref: & Vous devrez {\bf respecter scrupuleusement} les instructions de votre médecin . \\
Join: & Les instructions de votre médecin doivent être {\bf soigneusement surveillées} . \\
Feature: & Les instructions de votre médecin doivent être {\bf suivies attentivement} . \\
\hline
Src: & All injections of Macugen will be administered by your doctor. \\
Ref: & Toutes les injections de Macugen doivent être réalisées par votre médecin. \\
Join: & Toutes les injections de Macugen seront {\bf à l'ordre du jour} de votre médecin. \\
Feature: & Toutes les injections de Macugen seront {\bf effectuées} par votre médecin. \\
\hline
\end{tabular}
\end{center}
\caption{\label{tab:examples} Translation examples of in-domain medical sentences with and without domain feature.}
\end{table*}

\begin{table*}
\begin{center}
\begin{tabular}{|l|cccccc|}
\hline
\multirow{2}{*}{\bf Test} & \multicolumn{6}{c|}{\bf Domain feature}\\

    & IT & Literature & Medical & News & Parl. & Tourism\\
 \hline
IT             & \bf 54.56 & -12.76 & -10.25 & -12.43 & -13.83 & -14.18\\
Literature     & -5.96 & \bf 30.73 & -5.13 & -2.89 & -3.50 & -3.03\\
Medical        & -4.82 & -6.23 & \bf 42.51 & -5.06 & -5.39 & -4.74\\
News           & -3.36 & -1.58 & -3.04 & \bf 34.61 & -0.81 & -2.48\\
Parliamentary  & -4.14 & -1.92 & -3.09 & -0.39 & \bf 37.79 & -3.01\\
Tourism        & -6.72 & -3.2 & -4.16 & -4.26 & -4.35 & \bf 38.30\\
\hline
\end{tabular}
\end{center}
\caption{\label{tab:results2} BLEU score decreases using different predefined domain tags.}
\end{table*} 

Table \ref{tab:results} shows translation accuracy results for the different training configurations. Accuracies are measured using BLEU\footnote{\url{multi-bleu.perl}}.
As expected, the \textit{Join} model outperforms all \textit{Single} models on their corresponding test sets, showing that NMT engines benefit from additional training data. Differences in accuracy are lower for domains with a higher representation in the {\it Join} model, like Parliamentary and Tourism. No domain information is used on these first configurations ({\it none}).

Results for models incorporating domain information are detailed in columns \textit{Token} and \textit{Feature}. \textit{Oracle} experiments indicate that the test set domains are known in advance, thus allowing to use the correct side-constraint. The additional token approach gives mixed results; it improves translation quality on some tasks and degrades on some others compared to the \textit{Join} model. On the contrary, incorporating domain information through the {\it Feature} approach consistently improves translation quality on all the tasks. Adding domain information on all the source words seems to be a good technique to convey domain side-constraint and to improve NMT target words choice consistency. Differences between the {\it Feature} and {\it Join} configurations are shown in parentheses. 
Note that an average improvement of $0.80$ is observed on all test sets with the exception of Parliamentary translations, for which accuracy was only improved by $0.26$. This can be explained by the fact that Parliamentary is the best represented domain in {\it Join} training set.
 
Translation examples are shown in Table \ref{tab:examples} in a medical context. They show the impact on domain adaptation introduced by the {\it Feature} approach. The first example shows the preference of the {\it Feature} model for the French translation {\it suivies attentivement} of the English {\it carefully observed}. It seems more suitable than the hypothesis {\it soigneusement surveillées} output by the {\it Join} model. A similar effect is shown on the second example where the French {\it effectuées} is clearly more adapted as translation of {\it administered} than {\it à l'ordre du jour}.

Finally, we also evaluate the ability of our presented approach ({\it Feature}) to face test sets for which the domain is not known in advance. Hence, before translation, the domain tag is automatically detected using an in-house domain classification module based on RNN technology to disambiguate between the six different domains. The tool predicts the domain on a sentence-by-sentence basis, then translation is carried out using the predicted domain value in {\it Feature} model. Last row of Table \ref{tab:results} shows the accuracy of the domain classification tool for sentences on each of the predefined domains.

Results for this last condition are shown in column {\it RNN}. Event though domain is wrongly predicted in some cases, translation accuracy is still improved when compared to the {\it Join} model. Notice that domain classification at sentence level is a challenging task as short context is considered. We also confront our approach with a final test set from a brand new domain, {\it Dialogs}, that is not present in our training data. Sentences are selected from TED Talks corpora. The {\it RNN} toolkit is able to assign each test sentence to one of the source domains, leading to outperform the {\it Join} model.

In order to better understand the influence of the predicted domain, we conduct a final set of experiments. Using the {\it Feature} model, we run each test set using all domain values. Results are detailed in Table \ref{tab:results2} showing that translation quality can significantly be degraded when translating sentences with the wrong domain tag. It is especially the case for \textit{IT} domain, where translating with the wrong domain tag dramatically reduces accuracy. 
Results also reveal proximities between different domains like, for example, \textit{News} and \textit{Parliamentary}. Translating the \textit{News} test set with the \textit{Parliamentary} domain tag (and vice versa) does not seem to hurt translation quality compared to other domain tag mismatches. Inversely, when translation and reference are provided, using BLEU as a similarity measure we observe that the model learned to classify: this is actually expected as RNN and word embeddings provide more powerful discriminant features.

\section{Conclusions and Further Work}
\label{sec:conclusions}
We have presented a method that incorporates domain information into a neural network. It allows to perform domain-adapted translations using a unique network that covers multiple domains. The presented method does not need to re-estimate model parameters when performing translations on any of the available domains.

We plan to further improve the feature technique detailed in this work. Rather than providing the network with a hard decision about domain, we want to introduce a vector of distance values of the given source sentence to each domain, thus allowing to smooth the proximity of each sentence to each domain.

Additionally, Table \ref{tab:results2} shows indirectly that the neural network has learnt the ability to classify domains at the sentence level.  
We also plan to implement a joint approach for domain classification and translation, avoiding dependency with the RNN classifier.

Finally, since domain classification is a document level task, it would be interesting to extend the current study to document level translation.

\bibliography{acl2017}
\bibliographystyle{acl_natbib}

\end{document}